\newcommand{\CLASSINPUTinnersidemargin}{18mm}
\newcommand{\CLASSINPUToutersidemargin}{12mm}
\newcommand{\CLASSINPUTtoptextmargin}{20mm}
\newcommand{\CLASSINPUTbottomtextmargin}{25mm}
\documentclass[10pt,conference,a4paper]{IEEEtran}

\usepackage[utf8]{inputenc}

\usepackage{times}

\usepackage{graphicx}
\usepackage{subfigure}
\DeclareGraphicsExtensions{.png,.eps,.ps,.pdf}

\usepackage{url}

\begin{document}

\title{Improving Named Entity Recognition in Tor Darknet with Local Distance Neighbor Feature}

\author{
\IEEEauthorblockN{\small Mhd Wesam Al-Nabki}
\IEEEauthorblockA{\small Dept. IESA.\\
Universidad de Le\'{o}n\\
mnab@unileon.es}

\and

\IEEEauthorblockN{\small Francisco Jañez-Martino}
\IEEEauthorblockA{\small Dept. IESA.\\
Universidad de Le\'{o}n\\
fjanm@unileon.es}

\and

\IEEEauthorblockN{\small Roberto A. Vasco-Carofilis}
\IEEEauthorblockA{\small Dept. IESA.\\
Universidad de Le\'{o}n\\
rvasc@unileon.es}

\and

\IEEEauthorblockN{\small Eduardo Fidalgo}
\IEEEauthorblockA{\small Dept. IESA.\\
Universidad de Le\'{o}n\\
efidf@unileon.es}

\and

\IEEEauthorblockN{\small Javier Velasco-Mata}
\IEEEauthorblockA{\small Dept. IESA.\\
Universidad de Le\'{o}n\\
jvelm@unileon.es}
}

\maketitle

\begin{abstract}
Name entity recognition in noisy user-generated texts is a difficult task usually enhanced by incorporating an external resource of information, such as gazetteers. However, gazetteers are task-specific, and they are expensive to build and maintain. This paper adopts and improves the approach of Aguilar et al. by presenting a novel feature, called Local Distance Neighbor, which substitutes gazetteers. We tested the new approach on the W-NUT-2017 dataset, obtaining state-of-the-art results for the Group, Person and Product categories of Named Entities. Next, we added 851 manually labeled samples to the W-NUT-2017 dataset to account for named entities in the Tor Darknet related to weapons and drug selling. Finally, our proposal achieved an entity and surface F1 scores of 52.96\% and 50.57\% on this extended dataset, demonstrating its usefulness for Law Enforcement Agencies to detect named entities in the Tor hidden services.
\end{abstract}

\begin{IEEEkeywords}
NER, Darknet, gazetteer
\end{IEEEkeywords}

{\bf Type of contribution:}  {\it Published research}

\section{Introduction}

Named Entity Recognition (NER) is a cornerstone task in Natural Language Processing (NLP) systems focused in detecting Named Entities (NEs) in text inputs, i.e., if words refer to categories such as Persons or Places. NER can be used to identify people names and nicknames, shipping addresses or even references to groups or terrorist organizations in The Onion Router (Tor) network \cite{al2019darkner}, which is one of the most popular darknets that provides its users with a high level of anonymity and privacy. These features have made the Tor network a safe shelter for trading illegal products, such as drugs and weapons. Usually, NER approaches uses a gazetteer \cite{aguilar2017multi}, a fixed and expensive to maintain external resource of information. This paper is a summary of our previous work \cite{al2020improving}, where we proposed Local Distance Neighbor (LDN), a novel feature that substitutes the gazetteer.

\section{Background}

Several proposals have tackled the problem of recognizing textual entities from an input text. In 2017, the Workshop of Noisy User-generated Test (W-NUT) published the W-NUT-2017 with samples of noisy user-generated texts cropped from Twitter \cite{derczynski2017results}. This dataset was used by Lin et al. \cite{lin2017multi} to test a novel neural network that achieved an F1 score of  40.42\%, and Von D\"aniker et al. \cite{KADARI201831} improved that result with a Transfer Learning model that achieved an F1 of 40.78\%. Later, the model of Aguilar et al. \cite{aguilar2017multi}, which entirely depends on an external data resource, a gazetteer, obtained an entity and surface F1 scores of 41.86\% and 40.24\% respectively over the same dataset, and this model was surpassed on downstream tasks, such as NER by the novel approach of Akbik et al. \cite{akbik2019pooled} which scored an F1 score of 49.59\% over the W-NUT-2017.

\section{Methodology}

\begin{figure*}[t]
\centerline{
\includegraphics[width=\textwidth, height=9.2cm]{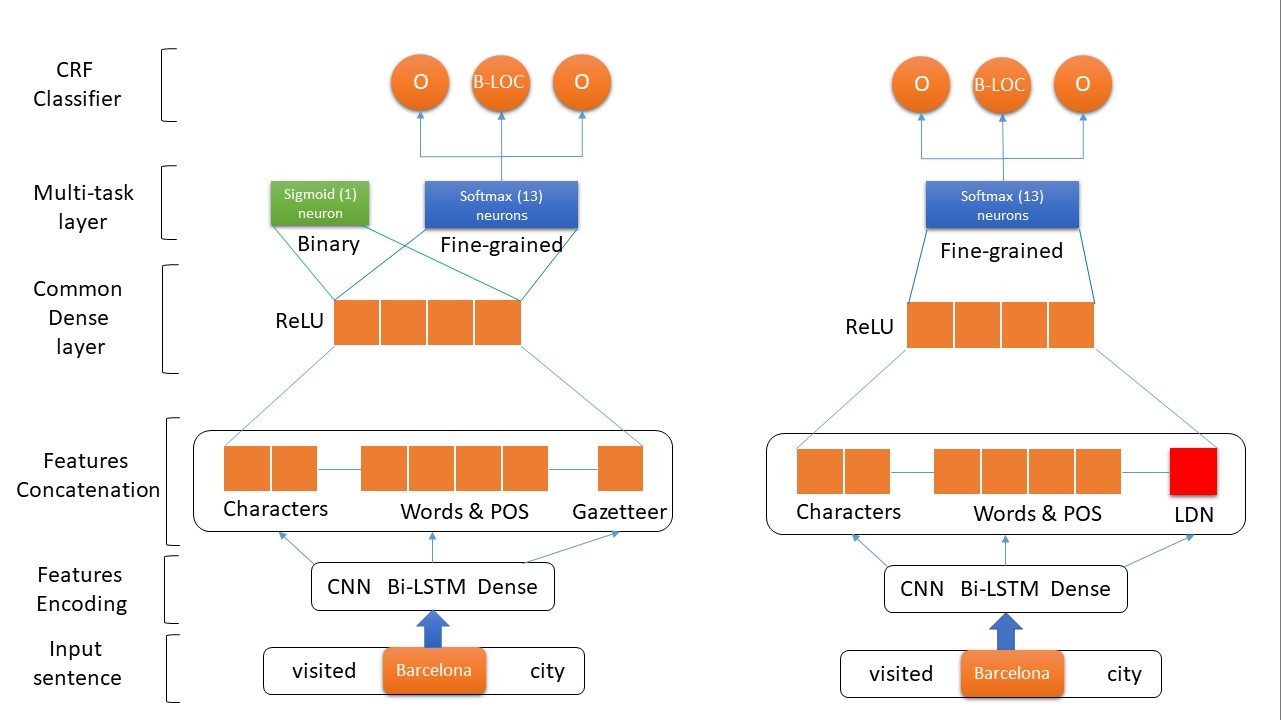}
}
\caption{Baseline model of Aguilar et al. \cite{aguilar2017multi} (left) and our proposed modification that introduced the \textit{LDN} \cite{al2020improving} and omitted the binary output (right).}
\label{fig:aguilar}
\end{figure*}

This work \cite{al2020improving} adopts the neural network architecture proposed by Aguilar et al. \cite{aguilar2017multi} and improves it by substituting its gazetteer with the LDN feature, as shown in Fig. \ref{fig:aguilar}. The model of Aguilar et al. involves three categories of features extracted from: (1) the \textit{characters} of the input token; (2) the context of the input \textit{word}; and (3) the presence in a \textit{lexicon}, which labels the tokens into the entity classes present in the gazetteer. Next, the character, word, and lexical vectors are passed into a multi-task network, that at the end gives way to a Conditional Random Field (CRF) to account for sequential constraints in the input text.

The proposed LDN feature is inspired in the way a human tries to determine the meaning of an unknown term inside a document, i.e., it uses the semantically-similar tokens of that term to predict its potential tags.

The algorithm of building the LDN feature consists of two phases. In the \textit{Initialization Phase}, each token in the training set is preprocessed, e.g. removing special characters and stop words, and embedded in an embedding space. The second phase is called the \textit{Accumulation Phase}, and it is triggered when a new query token is introduced. The algorithm uses the \textit{cosine similarity} to determine the $x$ most semantically-similar words to the query token, along with their tags, i.e. their categories. Based on the tags of these neighbors, the LDN algorithm determines the trend of the query token. For example, a query token is "Cordoba", and the majority of its neighbors were tagged as a location in the training set, like Barcelona and Madrid. Hence, there is a high probability that Cordoba is the name of a location. 

\section{Experimentation and Results}

The experiments were performed over the \textit{Noisy User-generated Text on Tor} (NUToT) dataset, which is an extended version of the W-NUT-2017 dataset that adds $851$ samples of the \textit{Weapon} and \textit{Drugs}\cite{NABKI2017} categories from the Darknet Usage Text Addresses (DUTA) \cite{ToRank_ESWA2019, al2017classifying} dataset. We considered $80\%$ of the dataset for training and $20\%$ for testing, and the three tested classifiers used $226$ training epochs.

We compared the models of Aguilar et al. \cite{aguilar2017multi} and Akbik et al. \cite{akbik2019pooled} with our proposal, whereas five neighbors were considered for each input token to estimate its LDN vector. The F1 Entity and Surface scores are shown in Table \ref{tab:results}, where our proposal obtained the best overall results (Total) in the Entity Score, outperforming Aguilar et al. \cite{aguilar2017multi} and Akbik et al. \cite{akbik2019pooled} in the Group, Person and Product tags of the F1 Entity scores.

\begin{table}[tbh]
\centering
\caption{Entity and Surface F1 scores of our proposal, and the models of Aguilar et al. \cite{aguilar2017multi} and Akbik et al. \cite{akbik2019pooled}}
\label{tab:results}

\begin{tabular}{lllllll}
\hline
\multicolumn{1}{c}{\textbf{Category}} & \multicolumn{2}{c}{\textbf{\begin{tabular}[c]{@{}c@{}}Aguilar et al. \\ model F1 (\%)\end{tabular}}} & \multicolumn{2}{c}{\textbf{\begin{tabular}[c]{@{}c@{}}Akbik et al.\\ model F1 (\%)\end{tabular}}} & \multicolumn{2}{c}{\textbf{\begin{tabular}[c]{@{}c@{}}Our Proposal)  \\ F1(\%)\end{tabular}}} \\ 
\hline
 & \textbf{Ent.} & \textbf{Surf.} & \textbf{Ent.} & \textbf{Surf.} & \textbf{Ent.} & \textbf{Surf.} 
\\ \cline{2-7}

\textbf{Corporation} & 18.96 & 19.46 & \textbf{29.36} & 31.91 & 21.92 & 20.94 \\
\textbf{Creative-work} & 18.86 & 21.35 & \textbf{26.58} & 23.14 & 22.83 & 24.31 \\
\textbf{Group} & 21.14 & 22.54 & 23.38 & 23.36 & \textbf{26.39} & 27.91 \\
\textbf{Location} & 35.06 & 33.10 & \textbf{60.43} & 61.00 & 54.58 & 54.87 \\
\textbf{Person} & 61.48 & 61.01 & 62.05 & 61.86 & \textbf{62.15} & 62.36 \\
\textbf{Product} & 53.25 & 53.53 & 53.64 & 51.95 & \textbf{63.15} & 61.57 \\ \hline
\textbf{Total} & 44.73 & 43.29 & 52.17 & 50.53 & \textbf{52.96} & 50.57 \\ \hline
\end{tabular}

\end{table}

\section{Conclusions}

In this paper we proposed a novel feature, called Local Distance Neighbor (LDN), to substitute gazetteer. We integrated the LDN feature with the model of Aguilar et al. \cite{aguilar2017multi} to replace the use of gazetteers. Using the NuToT dataset, we found that our approach outperforms the model of Aguilar et al. and three Named Entities categories: Product, People, and Group of the state of art solutions of Akbik et al. \cite{akbik2019pooled}. In the future, we are planning to incorporate graphical features extracted from images taken from Tor domains to improve the classification of Named Entities.

\section*{Acknowledgments}

This work was supported by the framework agreements between the Universidad de León and INCIBE (Spanish National Cybersecurity Institute) under Addendum $01$. 

This research has been funded with support from the European Commission under the 4NSEEK project with Grant Agreement 821966.

We acknowledge NVIDIA Corporation with the donation of the TITAN Xp and Tesla K40 GPUs used for this research.

\end{document}